\documentclass[11pt]{article}

\usepackage[utf8]{inputenc}   
\usepackage[T1]{fontenc}
\usepackage[margin=1in]{geometry}
\usepackage{amsmath,amssymb}
\usepackage{array,booktabs,multirow}
\usepackage{graphicx,xcolor}
\usepackage{pgfplots}
\pgfplotsset{compat=1.18}
\usepackage{microtype}
\usepackage[numbers,sort&compress]{natbib}  
\usepackage[hidelinks]{hyperref}
\usepackage{authblk}
\usepackage{float}
\usepackage{indentfirst}
\usepackage{titlesec}
\titleformat{\paragraph}[runin]{\normalfont\bfseries}{}{0pt}{}
\titlespacing{\paragraph}{\parindent}{\medskipamount}{1em}

\usepackage{newunicodechar}
\newunicodechar{—}{---}
\newunicodechar{§}{\S}
\newunicodechar{Δ}{\ensuremath{\Delta}}

\title{InfluMatch: Frontier-Quality KOL Search at 4B-Model Cost}
\author{Krittanon Kaewtawee, Petmongkon Pornpichitsuwan, Natchaya Temyingyong,\protect\\ Nutnicha Laplamoon, Wachiravit Modecrua,\protect\\ Krittin Pachtrachai, Touchapon Kraisingkorn}
\affil{\small Amity AI Holdings Co.,Ltd.}
\affil{\small\texttt{\{krittanon, petmongkon, natchaya, nutnicha.lap, wachiravit, krittin, touchapon\}@amity.co}}
\date{}

\begin{document}
\maketitle
\begin{abstract}
Matching influencers (KOLs) to free-form, multi-part Thai marketing criteria is today served either by keyword search over structured profiles,
which misses semantic fit, or by prompting frontier LLMs over every candidate, which is accurate but slow and expensive.
We present InfluMatch, a low-cost three-stage cascade — retrieval $\rightarrow$ rerank $\rightarrow$ reason 
— built entirely from small open-weight models: dense retrieval returns 50 candidates, 
a 4B pointwise reranker scores each by the log-probability of a single Yes token and keeps 10, 
and a 4B reasoner grades the shortlist per criterion on a rubric with a Thai rationale.
The cascade is designed for cost: reasoning over a filtered top-10 halves token spend versus reasoning over all 50 while scoring 14 points higher. 
End-to-end against human relevance labels on an 11-query set with all 50 candidates labeled, the full cascade reaches \textbf{94.1\%} P@5, 
versus a retrieval-only baseline near random; it matches the frontier model Kimi-K2.6 (91.8\%) while emitting ${\sim}35\times$ fewer output tokens and serving a 50-KOL query in ${\sim}20$\,s 
on one A100. Notably, the only fine-tuning that pays off is pairwise: a SimPO-tuned reranker matches the frontier baseline's best-pick accuracy (78.0 EM), 
whereas fine-tuning the reasoner on pointwise per-criterion labels improves offline scores yet degrades end-to-end ranking 
— an inversion we trace to the design of the absolute labeling task — 
leaving the untuned base model as the strongest deployed reasoner. The result is a deployable, explainable KOL search system at a small fraction of frontier serving cost.
\end{abstract}

\section{Introduction}
\label{sec:introduction}

Influencer marketing allocates a large share of brand budgets, and its central
operational problem is \emph{matching}: given a campaign, find the Key Opinion
Leaders (KOLs) whose audience, content, and voice fit the brief. In the Thai
market, a marketer's intent is naturally expressed as free-form, multi-part
criteria --- ``a female food reviewer with a playful tone whose audience skews
young'' --- while the supply side is a long tail of creators whose relevance
must be inferred from heterogeneous, largely Thai signals (bios, video
transcripts, audience profiles) rather than read off a clean attribute table.

\paragraph{Why keyword and structured search fall short.}
The incumbent approach treats matching as lookup over a scraped KOL database
filtered by keywords and hard attributes. This fails on three counts: it is
\emph{lexical, not semantic} (a ``playful food reviewer'' query misses a KOL
whose bio never says so but whose content plainly fits); the signals are
\emph{fragmented} across profile fields, transcripts, and engagement metrics,
so a query can satisfy each field in isolation yet return a poor overall match;
and the deciding criteria are \emph{dynamic}, campaign-specific predicates that
a static schema cannot anticipate. Our evaluation confirms the ceiling of pure
recall: dense retrieval alone barely separates from a random baseline on
end-to-end precision (54.5\% vs.\ 54.0\% P@5, \S\ref{sec:results-e2e}).

\paragraph{Approach.}
We present \textbf{InfluMatch}, a three-stage pipeline --- \emph{retrieval}
$\rightarrow$ \emph{rerank} $\rightarrow$ \emph{reason} --- that turns marketer
criteria into a ranked KOL shortlist with per-criterion scores and Thai
rationales. Stage~1 retrieves a top-50 candidate set from a vector index;
Stage~2 applies a pointwise 4B LLM reranker that keeps the top~10; Stage~3
scores each shortlisted KOL per criterion and combines the scores by an even
sum. To supervise the learned stages we build an evidence-grounded,
multi-objective human-labeled corpus over synthetic-but-realistic Thai briefs
--- pointwise ordinal scores, binary pass/fail verdicts, and pairwise
best/worst judgments --- and fine-tune two 4B open-weight students: a pairwise SimPO reranker and a
pointwise SFT$+$GRPO reason scorer. Only the pairwise student survives
end-to-end evaluation (\S\ref{sec:discussion}); the deployed reasoner is the
untuned base model.

\subsection{Contributions}
\label{sec:contributions}

\begin{itemize}
  \item \textbf{A deployable three-stage KOL matching pipeline} for free-form
  Thai criteria, returning a ranked shortlist with per-criterion scores and
  Thai rationales. End-to-end, the full cascade lifts P@5 from
  $\approx$54.5\% (retrieval only) to \textbf{94.1\%} on an 11-query pool
  with all 50 candidates labeled, while emitting ${\sim}35\times$ fewer
  output tokens than a frontier model.
  \item \textbf{A synthetic brief-generation pipeline and a multi-objective,
  evidence-grounded labeled corpus} of pointwise ordinal scores, binary
  pass/fail verdicts, and pairwise best/worst judgments over realistic Thai
  briefs.
\item \textbf{A pairwise SimPO reranker (4B)} that matches the frontier
  baseline's best-pick accuracy (78.0 EM) and transfers downstream,
  improving end-to-end P@5 over the untuned reranker.
  \item \textbf{An analysis of an offline--end-to-end inversion}: pointwise
  SFT$+$GRPO fine-tuning of the reason scorer wins on per-criterion metrics
  but loses end-to-end to its untuned base; an evidence audit traces the gap
  to the design of the absolute labeling task, indicating that the recoverable
  supervision signal lives in relative judgments.
\end{itemize}
\section{Background and Related Work}
\label{sec:related}

\paragraph{Influencer--brand matching.}
Industrial KOL discovery is dominated by database platforms that filter
scraped creator profiles by keywords and hard attributes and surface
candidates for manual review. Academic treatments formalize the problem as
two-sided matching or recommendation over structured features, and
marketing-science work models the market-level consequences of AI-assisted
matching.
Both lines presuppose that campaign intent is already encoded in a fixed
attribute schema; neither addresses free-form, multi-part briefs whose
deciding predicates are campaign-specific, nor produces per-criterion,
evidence-grounded justifications a marketer can audit. We instead cast
matching as criteria-conditioned retrieval and grading over heterogeneous,
largely Thai creator evidence.

\paragraph{Multi-stage cascades for ranked retrieval.}
Telescoping retrieval through stages of increasing per-candidate cost is a
classical effectiveness--efficiency design and remains the dominant shape of
neural IR: an efficient dense or lexical first stage
\citep{karpukhin2020dpr} feeds a stronger
cross-encoder reranker \citep{nogueira2019passage}, optionally followed by a
pairwise final stage \citep{nogueira2019multistage};
multilingual embeddings extend dense first stages to Thai.
InfluMatch instantiates this cascade with an explicit per-stage token and
latency budget, but departs at the last stage: rather than emitting only a
permutation, the reason stage grades each criterion on a rubric
with a Thai rationale, so the output is an auditable decision rather than an
opaque ordering.

\paragraph{LLMs as rerankers.}
Prompted LLM rerankers span pointwise, pairwise \citep{qin2024prp}, and
listwise families.
Pointwise relevance generation scores a candidate by the likelihood of a
relevance token \citep{nogueira2020monot5}; listwise prompting, e.g.,
RankGPT \citep{sun2023rankgpt}, achieves the strongest
benchmark quality but pays with long prompts, sliding-window passes, and
latency. A parallel line trains open-weight students, typically 7B listwise
rerankers, on proprietary teacher orderings
\citep{pradeep2023rankvicuna,pradeep2023rankzephyr}.
Our reranker sits deliberately at the cheap end of this space: pointwise
single-token scoring is embarrassingly parallel and emits $O(1)$ output
tokens per candidate, fitting a production latency budget that listwise
sliding windows do not. Unlike prior pointwise students trained with
cross-entropy on absolute relevance labels, we train the 4B student
on \emph{human pairwise preferences} with SimPO, which
\S\ref{sec:discussion} shows is the distinction that matters end-to-end.

\paragraph{LLM judges and rubric-based grading.}
Our reason stage is closer to LLM-as-a-judge than to reranking: it grades a
(criterion, KOL) pair against a rubric and must justify the grade. Judging
with strong proprietary LLMs is standard practice \citep{zheng2023judging},
and open evaluators such
as Prometheus \citep{kim2024prometheus} show that rubric-following judgment
can be distilled into
mid-sized open-weight models. This literature also documents biases and
calibration problems of absolute scoring relative to comparative judgment
\citep{wang2024fair};
our deployment
adds end-to-end evidence in the same direction, with the untuned base model
remaining the strongest deployed judge (\S\ref{sec:discussion}).

\paragraph{Post-training and the choice of supervision signal.}
Offline preference optimization aligns a policy directly on comparisons
\citep{rafailov2023dpo} ---
notably SimPO, a reference-free method whose implicit reward is the
length-normalized sequence log-probability \citep{meng2024simpo} --- while
Group Relative Policy Optimization (GRPO) optimizes against verifiable
rewards without a critic \citep{shao2024deepseekmath}.
Orthogonal to the algorithm is which \emph{label geometry} to collect:
relative judgments (best--worst scaling) are markedly more reliable than
rating scales at equal annotation budget \citep{kiritchenko2017bws},
and RLHF pipelines standardized on pairwise comparisons partly for this
reason; offline gains failing to survive
deployment is itself a documented pattern in recommender-system evaluation
\citep{garcin2014offline}. Our offline--end-to-end inversion is
consistent with this literature at deployment scale: from the same corpus,
the pairwise SimPO student transfers downstream while the pointwise
SFT$+$GRPO student does not.

\section{Method}
\label{sec:method}

\subsection{Problem Formulation}
\label{sec:formulation}

Given a marketer brief expanded into a set of matching criteria
$C = \{c_1, \dots, c_m\}$ and a candidate pool of KOLs
$\mathcal{K} = \{k_1, \dots, k_N\}$, each represented by heterogeneous Thai
signals, the task is to produce a ranked shortlist $\pi$ over $\mathcal{K}$
such that KOLs satisfying $C$ rank highest, together with a per-criterion
score and a Thai rationale for each shortlisted candidate.

\subsection{Cascade Overview}
\label{sec:pipeline}

\begin{figure}[H]
\centering
\includegraphics[width=\linewidth]{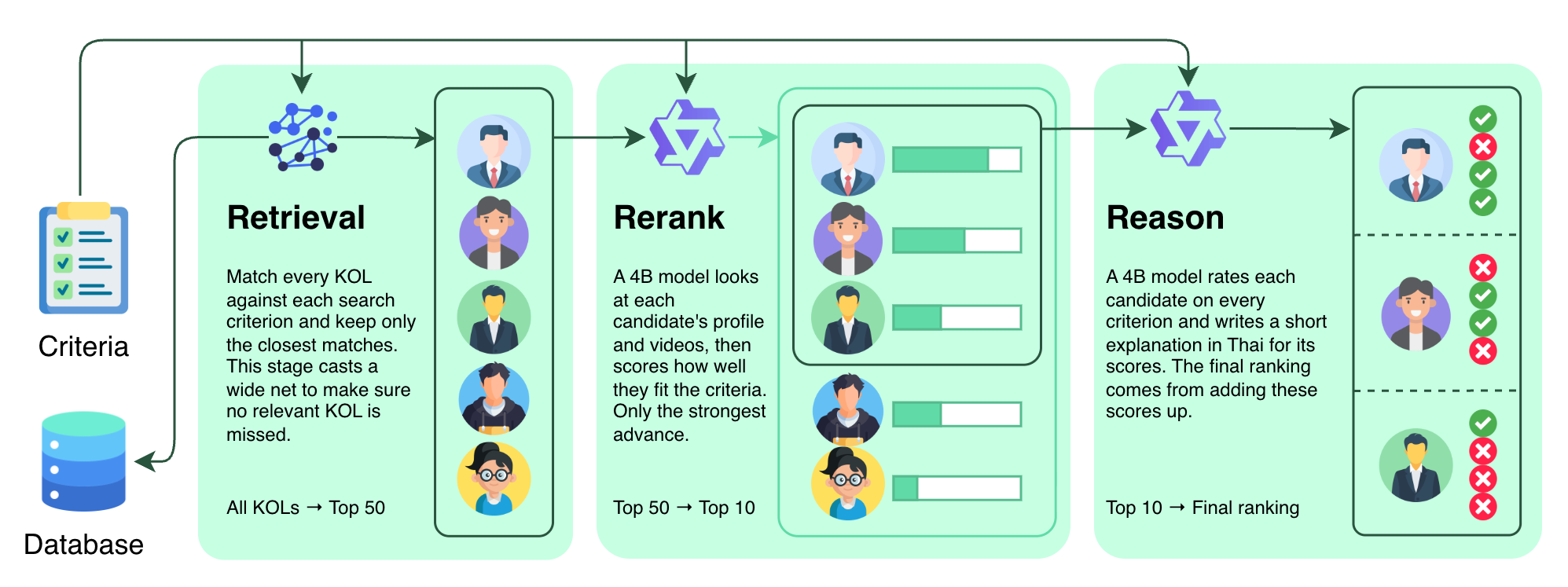}
\caption{The three-stage InfluMatch pipeline: retrieval $\rightarrow$ rerank
$\rightarrow$ reason. Each stage narrows the candidate set before the final
per-criterion scoring.}
\label{fig:3stage}
\end{figure}

InfluMatch decomposes matching into three stages of increasing
per-candidate cost, so that the expensive reasoning stage runs only on a
filtered shortlist (Figure~\ref{fig:3stage}, Table~\ref{tab:stages}).

% ---------------------------------------------------------------------------
% FIGURE — three-stage pipeline
% ---------------------------------------------------------------------------

% ---------------------------------------------------------------------------
% TABLE — Stage summary
% ---------------------------------------------------------------------------
\begin{table}[H]
\centering
\caption{The three pipeline stages. ``Out'' is the candidate count each stage
emits.}
\label{tab:stages}
\small
\begin{tabular}{@{}llll@{}}
\toprule
Stage & Mechanism & In~$\rightarrow$~Out & Output \\
\midrule
Retrieval & dense retrieval & corpus~$\rightarrow$~50 & ranked candidates \\
Rerank & pointwise scorer, $\log p(\texttt{Yes})$ & 50~$\rightarrow$~10 & relevance scores \\
Reason & per-criterion $\{0,1,2\}$ scoring, parallel & 10~$\rightarrow$~final & scores $+$ Thai rationale \\
\bottomrule
\end{tabular}
\end{table}

\paragraph{Stage 1: Retrieval.}
Each criterion is embedded as its own query vector and searched against the
pooled KOL vector table. A KOL's similarity to a criterion is its
\emph{best-matching} vector --- a per-KOL \texttt{MIN}-distance (least cosine
distance) over all of that KOL's pooled vectors --- and the per-criterion
similarities are summed into a single retrieval score per KOL. KOLs are ranked by
this score, an optional hard \texttt{tier} filter is applied, and the top~50
candidates are kept, with their profile and top videos fetched for the
downstream stages. This dense-retrieval stage is tuned for \emph{recall} --- not losing
relevant KOLs from the candidate pool --- rather than final ordering; the
ordering work is done downstream by the rerank and reason stages.

\paragraph{Stage 2: Rerank.}
A 4B pointwise reranker scores each candidate in a single forward pass: the
model receives the criteria together with the candidate's profile details and
video summaries, and the relevance score is the log-probability of a single
\texttt{Yes} token. The top-10 candidates advance. Emitting one scored token
per candidate keeps the stage's output cost at ${\sim}0.1$k tokens per query.

\paragraph{Stage 3: Reason.}
A 4B reason scorer grades each shortlisted KOL against each criterion $c_i$
on a $\{0,1,2\}$ ordinal rubric and produces a Thai rationale. The final
ranking score is the even sum $s(k) = \sum_{i=1}^{m} s_i(k)$.
The deployed scorer is the untuned base model (\S\ref{sec:discussion}); the
fine-tuned variant below is the experimental arm whose end-to-end behavior
we analyze.

\subsection{Supervision Signals}
\label{sec:supervision}

We collect three label types over (criteria, KOL) units drawn from
synthetic-but-realistic Thai briefs (\S\ref{sec:data-creation}):
(i)~\emph{pointwise} per-criterion ordinal scores in $\{0,1,2\}$;
(ii)~KOL-level \emph{binary} pass/fail verdicts; and (iii)~\emph{pairwise}
best/worst judgments over candidate groups. The reranker learns from
relative judgments, the reason scorer from absolute ones.

\subsection{Reranker Training: Pairwise SimPO}
\label{sec:reranker-training}

We fine-tune the reranker, as a LoRA adapter \citep{hu2022lora}, with SimPO
\citep{meng2024simpo} on preference
pairs $(k_w, k_l)$ of a preferred and dispreferred candidate under the same
criteria. Because the scored response is a single token, SimPO's
length-normalized reward reduces to the \texttt{Yes} log-probability itself,
and the objective becomes a margin-augmented logistic loss on the score
difference:
\begin{equation}
\mathcal{L} \;=\; -\log \sigma\!\big(\beta\,(s_w - s_l) - \gamma\big),
\qquad s = \log p(\texttt{Yes} \mid C, k),
\label{eq:simpo}
\end{equation}
where $p(\texttt{Yes}\mid C,k)$ is the reranker's probability of the
\texttt{Yes} token given criteria $C$ and candidate $k$ --- so $s_w$ and $s_l$
are this score at the preferred and dispreferred candidate $k_w,k_l$ --- $\sigma$
is the logistic sigmoid, $\beta$ an inverse temperature, and $\gamma$ a fixed
target margin. This is reference-free and directly shapes
the quantity used at inference, so training and serving optimize the same
score.

\paragraph{SFT-judge baseline.}
As a fine-tuning control, we also train a \emph{direct} A-vs-B judge
(SFT-judge) by supervised fine-tuning on the same pairwise labels: the model
receives the criteria and two candidates and is trained to emit the token of
the preferred one (\texttt{A} or \texttt{B}). SFT-judge and SimPO thus share
the training data and differ only in objective and inference path ---
generative A-vs-B judgment versus pointwise $\log p(\texttt{Yes})$ scoring
--- isolating the contribution of the preference objective
(\S\ref{sec:results-rerank}).

\subsection{Reason Scorer Training: Pointwise SFT+GRPO}
\label{sec:reasoner-training}

To test whether absolute per-criterion supervision transfers to end-to-end
ranking, we fine-tune the reason scorer in two phases on human pointwise
labels; the labels contain ordinal scores only, with no reference rationales.
\textbf{SFT} adapts the model to the per-criterion scoring format.
\textbf{GRPO} \citep{shao2024deepseekmath} then optimizes against the human
labels with a single gated
score reward: a completion earns non-zero reward only if it passes a joint
gate --- a valid ordinal score and a non-empty rationale --- and gated
completions are paid by ordinal distance to the human label: $1.0$ for an
exact match, $0.2$ for an off-by-one prediction, and $0$ for a flipped one.
The rationale is thus never directly supervised; the gate only requires that
one be produced.

\section{Data Creation}
\label{sec:data-creation}

Our corpora are produced by a three-stage pipeline: (i)~\emph{synthetic brief
generation} fabricates realistic, brand-grounded Thai marketer briefs;
(ii)~\emph{criteria synthesis} expands each brief into five matching criteria;
and (iii)~\emph{human annotation} collects pointwise, binary, and pairwise
judgments over (brief, criteria, KOL) units.
Tables~\ref{tab:pipeline-stats} and~\ref{tab:label-stats} summarize the
generation pipeline and the resulting labeled datasets.
\subsection{Synthetic Briefs and Criteria}
\label{sec:synth-brief}

Real marketer briefs are scarce and proprietary, so we synthesize briefs
mirroring the schema of real reference Thai campaign briefs. Each brief is
seeded by a consumer and a marketer persona sampled from
Nemotron-Personas-USA \citep{nvidia2025nemotronpersonas}; a campaign scenario
is drawn from a six-axis taxonomy
under non-uniform priors, constrained by 13 hard rules that forbid incoherent
combinations (e.g.\ a crisis-recovery objective forces a reactive context).
The campaign subject is grounded with a live web search, and the brief text
--- a structured object with brand, product, objective, audience, and
tone fields --- is written by
\texttt{DeepSeek-V3.2} \citep{deepseek2025v32}, with marketer expertise
sampled over three levels to control jargon.

Each brief is then expanded into exactly \textbf{five} Thai criteria by a
single \texttt{DeepSeek-V3.2} call, aligned to five fixed dimensions:
(C1)~audience fit, (C2)~content--product relevance, (C3)~selling and
persuasion style, (C4)~communication credibility, and (C5)~content format and
execution. The criteria are \emph{dual-purpose}: they serve as the Stage-1
retrieval queries and as the labeling rubric shown to annotators.

\subsection{Human Annotation}
\label{sec:human-annotation}

\begin{figure}[H]
\centering
\includegraphics[width=\linewidth]{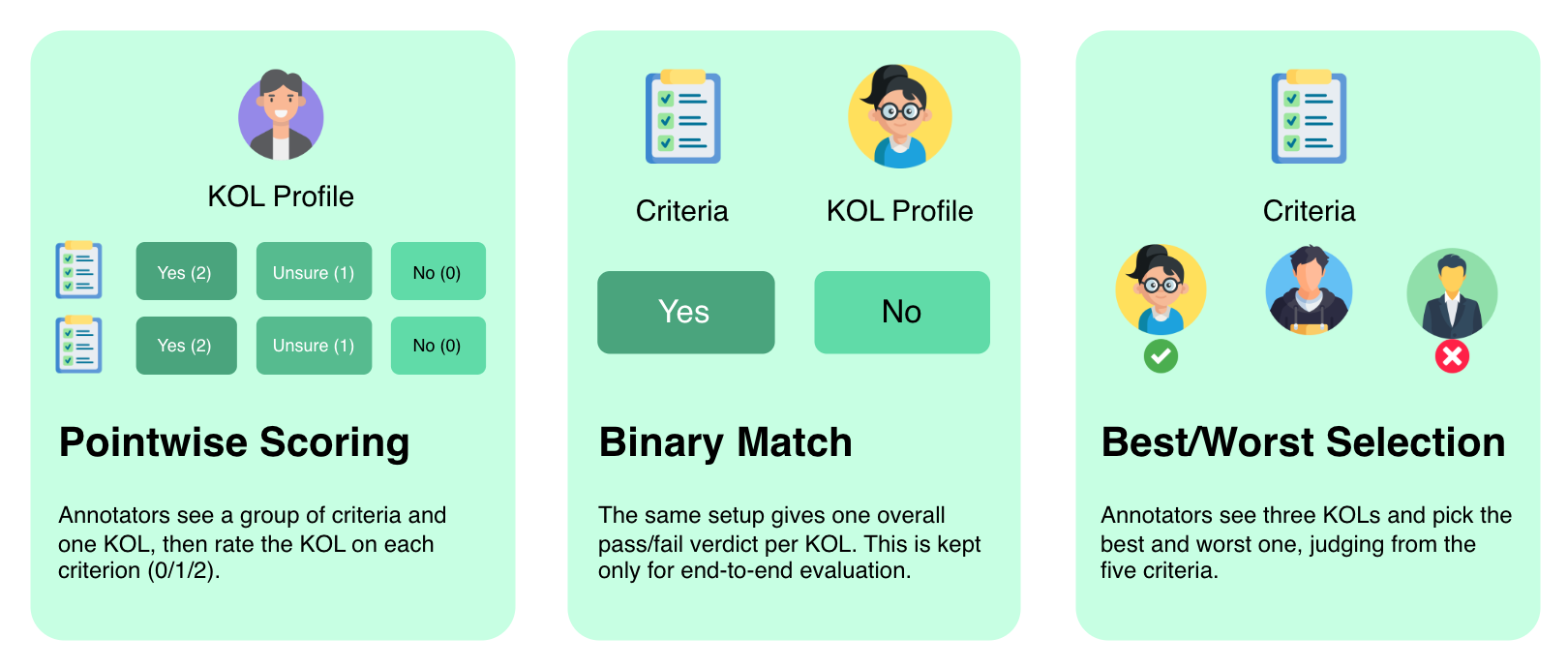}
\caption{The three annotation interfaces presented to human annotators:
pointwise scoring (T2), where each criterion receives an ordinal
$0/1/2$ score; binary match (E2E), which elicits one overall pass/fail
verdict per KOL for end-to-end evaluation; and best/worst selection
(T10), where annotators pick the best and worst of three KOLs.}
\label{fig:annotation-tasks}
\end{figure}

Human supervision is collected through a dedicated labeling application over
three task types (Table~\ref{tab:label-stats}). \textbf{T2} shows the
criteria and one KOL, and elicits an ordinal score $\in\{0,1,2\}$ per
criterion plus a KOL-level binary pass/fail verdict; the latter is reserved
for end-to-end evaluation. Annotators judge each criterion directly: no
scoring rubric defining the $0/1/2$ boundaries was provided, leaving the
threshold between scores to each annotator's judgment. The interface also
presents items grouped by criteria set, so an annotator scores all candidates
under one brief consecutively. \textbf{T10} shows three KOLs and elicits a
best/worst pick, conditioning on the five criteria.

\subsection{Label Quality}
\label{sec:label-quality}

\paragraph{Evidence audit (T2).}
A Kimi audit of \textbf{873} human (KOL, criterion) pairs checks whether each
score is backed by \emph{verbatim} profile quotes, each verified as an exact
substring of the profile --- a model-independent grounding check that bounds
the label ceiling any model can reach (full treatment in Results). Grounding
is starkly score-dependent (Table~\ref{tab:evidence-audit}): only
\textbf{15.9\%} of score-2 (``match'') labels are evidence-supported. Since
score-2 is the largest class ($\approx$48\%) yet $\approx$84\% ungrounded, an
evidence-faithful model \emph{must} disagree with many human labels --- the
central finding developed in Results.

\begin{table}[t]
\centering
\caption{T2 Kimi evidence audit}
\label{tab:evidence-audit}
\small
\begin{tabular}{@{}lrr@{}}
\toprule
Human score & Pairs & Evidence-supported (\%) \\
\midrule
0 & 337 & 78.9 \\
1 & 115 & 54.8 \\
2 & 421 & 15.9 \\
\bottomrule
\end{tabular}
\end{table}

\begin{table}[t]
\centering
\caption{Synthetic generation pipeline.}
\label{tab:pipeline-stats}
\small
\begin{tabular}{@{}lll@{}}
\toprule
Generative step & Input & Output \\
\midrule
Problem inference & Campaign, Persona & Problem \\
Subject grounding & Problem, Campaign & Evidence \\
Brief generation & Problem, Evidence, Persona, Constraints & Brief \\
Criteria synthesis & Brief & 5 criteria \\
\bottomrule
\end{tabular}
\end{table}

\begin{table}[t]
\centering
\caption{Labeled datasets.}
\label{tab:label-stats}
\small
\begin{tabular}{@{}llll@{}}
\toprule
Task ID & Task & Input & Label type \\
\midrule
T2 & Pointwise scoring & criterion $+$ KOL & ordinal $0/1/2$ \\
E2E & Binary match & criteria $+$ KOL & pass / fail \\
T10 & Best/worst selection & 5 criteria $+$ 3 KOLs & best / worst \\
\bottomrule
\end{tabular}
\end{table}
\section{Experimental Setup}
\label{sec:setup}

We evaluate at two levels: \emph{offline}, testing each learned stage against
its own held-out human labels, and \emph{end-to-end}, testing the full cascade
against human relevance labels while measuring serving cost.

\paragraph{Model arms.}
Throughout, \textbf{Base} denotes the untuned \texttt{Qwen3.5-4B}
\citep{qwen2026qwen35};
\textbf{SFT} and \textbf{SFT$+$GRPO} denote the fine-tuned reason scorers of
\S\ref{sec:reasoner-training}; \textbf{SimPO} denotes the fine-tuned reranker
of \S\ref{sec:reranker-training}; and \textbf{Kimi-K2.6}
\citep{moonshot2026kimik26} is the frontier
baseline, run under the same prompts and input format as the students. The
deployed configuration is SimPO (text) rerank followed by Base reason.

\subsection{Offline Evaluation: Reranker}
\label{sec:setup-rerank}

\paragraph{Data.}
The human pairwise best/worst labels on the T10 test split
(\S\ref{sec:data-creation}).

\paragraph{Metric: best-pick Exact Match (EM).}
For each triplet the model compares the human-annotated \emph{best} and
\emph{worst} KOL (the middle KOL is ignored); the prediction is correct iff
it ranks the best above the worst, and EM is the fraction of triplets decided
correctly. The two training objectives induce two distinct inference paths,
reported side by side: a \emph{direct} generative A-vs-B judge (used by the
frontier baseline and the SFT student) and an \emph{indirect}
$\log p(\texttt{Yes})$ scorer (used by Base and the SimPO student); their
prompt formats are shown in Table~\ref{tab:rerank-prompts}. For the indirect
scorer we also report the \emph{margin}, the mean separation
$\overline{\,\log p(\texttt{Yes}\mid\text{best}) -
\log p(\texttt{Yes}\mid\text{worst})\,}$ in nats --- a calibration signal
orthogonal to EM.

\paragraph{Baselines.}
(i)~a \textbf{random} baseline (50\% EM by construction, since each triplet
is a binary best-vs-worst decision); (ii)~the \textbf{Kimi-K2.6} frontier
baseline as a direct A-vs-B judge; and (iii)~\textbf{Base} under both
inference paths --- the direct judge and the indirect
$\log p(\texttt{Yes})$ scorer.

\begin{table}[t]
\centering
\small
\caption{Minimal reranker prompt formats. The \emph{text$+$vision} variant
also attaches each video's cover thumbnail, which \emph{text-only} drops.}
\label{tab:rerank-prompts}
\begin{tabular}{@{}p{0.17\linewidth}p{0.76\linewidth}@{}}
\toprule
Path & Prompt \\
\midrule
Indirect \newline (score 1 KOL) &
\emph{System:} rate whether \textsc{one} influencer is a strong overall match
for the criteria. \newline
\emph{Input:} criteria (1--5) $+$ one KOL (profile summary $+$ up to 10 video
summaries $+$ thumbnails in text$+$vision) $+$ ``Answer \texttt{Yes} or
\texttt{No}.'' \newline
\emph{Output:} \texttt{Yes}/\texttt{No} \\
\addlinespace
Direct \newline (judge 2 KOLs) &
\emph{System:} decide which of \textsc{two} influencers better satisfies the
criteria, using only the provided evidence. \newline
\emph{Input:} criteria (1--5) $+$ KOL~A $+$ KOL~B, each serialized as above
(text-only or $+$thumbnails). \newline
\emph{Output:} \texttt{A} or \texttt{B} \\
\bottomrule
\end{tabular}
\end{table}

\subsection{Offline Evaluation: Reason Scorer}
\label{sec:setup-reason}

\paragraph{Data.}
The T2 pointwise corpus (\S\ref{sec:data-creation}): (KOL, criterion-set)
units split into train and test, each yielding five criterion-level ordinal
labels. All models are evaluated against held-out human labels.

\paragraph{Metric: support-weighted F1 (wf1).}
We report support-weighted F1 over the three ordinal classes. We weight by
support rather than macro-averaging because the class distribution is heavily
skewed (score-2 is $\approx$48\% of labels, \S\ref{sec:data-creation}) and
the deployed product cares about the dominant ``yes'' class, which macro-F1
under-weights. The distinction is not cosmetic: the same frontier-baseline
prompt reads as 33.6 macro-F1 but 38.4 weighted-F1, because its class-1 F1
is 0.00 (it never emits a confident ``unsure'').

\paragraph{Baselines.}
(i)~a stratified \textbf{random} baseline that samples from the gold class
distribution (20-seed average); (ii)~the \textbf{Kimi-K2.6} frontier
baseline; and (iii)~\textbf{Base}, prompted for score-only output.

\subsection{End-to-End Evaluation}
\label{sec:setup-e2e}

\paragraph{Query sets.}
Two human-labeled query sets are drawn from the binary (pass/fail)
annotations:
\begin{itemize}
  \item \textbf{Set~1} --- 11 queries, each with all 50 retrieved KOLs
  labeled, so precision is measured over the complete re-ranked list;
  \item \textbf{Set~2} --- 31 queries, each with the top-10 KOLs labeled.
\end{itemize}

\paragraph{Protocol.}
Retrieval is held fixed: the 50-candidate set per query is frozen, and only the
rerank and reason stages are re-run with different models. All models use the
same prompts and input format as the production pipeline, so measured numbers
reflect the deployed system rather than an idealized re-implementation.
Unless stated, reason runs on the rerank top-10; reason applied directly to
all 50 candidates (no rerank) is reported as a contrast.

\paragraph{Relevance and metrics.}
A KOL is relevant under \emph{binary majority}: with vote fraction
$f = \text{pass}/\text{total}$, it is positive if $f > 0.5$, negative if
$f < 0.5$, and dropped if $f = 0.5$. Over the relevant set of size $R$ per
query we report, at cutoff $k{=}5$: $\text{P}@k = \text{hits}@k / \min(k,R)$;
$\text{MAP}_\text{B}@k$, binary-relevance average precision normalized by
$\min(R,k)$; and $\text{nDCG}@k = \text{DCG}@k / \text{IDCG}(\min(k,R))$.
All metrics are macro-averaged over queries with $R \ge 1$.
\section{Results}
\label{sec:results}

\subsection{Reranker: Offline Best-Pick Accuracy}
\label{sec:results-rerank}

\begin{table}[H]
\centering
\caption{Reranker best-pick EM on the human pairwise T10 test set.}
\label{tab:rerank-em}
\begin{tabular}{lccc}
\toprule
 & \multicolumn{2}{c}{EM (\%)} & Margin \\
\cmidrule(lr){2-3}
Model & Direct judge & $\log p(\texttt{Yes})$ & (nats) \\
\midrule
Random        & 50.0          & 50.0          & ---     \\
Kimi-K2.6     & 78.0 (32/41)  & ---           & ---     \\
Base          & 75.6 (31/41)  & 63.4 (26/41)  & $+0.76$ \\
SFT           & 75.6 (31/41)  & ---           & ---     \\
SimPO         & ---           & \textbf{78.0} (32/41) & $+4.28$ \\
\bottomrule
\end{tabular}
\end{table}

Table~\ref{tab:rerank-em} reads down each column, since the two inference
paths are not strictly comparable cell-for-cell. On the indirect
$\log p(\texttt{Yes})$ path --- the one the pipeline serves --- SimPO training
lifts Base from 63.4 to \textbf{78.0} EM ($+14.6$ points) and widens the
best--worst margin roughly $6\times$ ($+0.76 \to +4.28$). The resulting 4B
student matches the frontier baseline's direct-judge EM (78.0). On the direct
path, the judge-style SFT (75.6) sits at Base's level: the gain is
attributable to the SimPO objective, not the SFT-judge format.

\subsection{Reason Scorer: Offline Per-Criterion Accuracy}
\label{sec:results-reason}

\begin{table}[H]
\centering
\caption{Reason-scorer offline leaderboard (support-weighted F1 against
held-out human labels).}
\label{tab:reason-leaderboard}
\small
\begin{tabular}{@{}llccr@{}}
\toprule
Model & Eval mode & wf1 & $\Delta$ vs no-think & $n$ \\
\midrule
Random baseline & --       & 43.6 & --     & 259 \\
Kimi-K2.6       & no-think & 36.8 & --     & 250 \\
Kimi-K2.6       & think    & 38.4 & --     & 244 \\
\midrule
Base            & no-think & 52.9 & --     & 254 \\
Base            & think    & 44.9 & $-8.0$ & 254 \\
SFT             & no-think & 57.7 & --     & 244 \\
SFT$+$GRPO      & no-think & 55.9 & --     & 254 \\
SFT$+$GRPO      & think    & \textbf{59.0} & $+3.1$ & 254 \\
\bottomrule
\end{tabular}
\end{table}

Table~\ref{tab:reason-leaderboard} reports the offline leaderboard. The
frontier baseline scores \textbf{38.4} wf1 --- \emph{below} the stratified
random floor of 43.6 --- because its evidence-faithful behavior collapses the
middle class and withholds the generous ``yes'' that human annotators assign
freely. Base reaches \textbf{52.9}, above both. Fine-tuning helps offline:
score-only SFT reaches \textbf{57.7} ($+19.3$pp over the frontier baseline,
$+14.1$pp over random), and SFT$+$GRPO tops the leaderboard at \textbf{59.0}
wf1 in think mode. During GRPO, training reward climbs from 0.5 to 0.94 over
75 steps; since the gate pays only completions containing a rationale, the
reward curve also certifies that the model learns to explain its scores.

\paragraph{Chain-of-thought.}
The effect of \emph{think} mode \citep{wei2022cot} depends on training:
thinking degrades Base
by 8.0pp (52.9 $\to$ 44.9) but helps SFT$+$GRPO by 3.1pp (55.9 $\to$ 59.0).

\subsection{End-to-End Ranking}
\label{sec:results-e2e}

\paragraph{Set~1 (full 50-KOL labeled pool).}
Table~\ref{tab:e2e-set1} reports Set~1 ($n{=}11$). Retrieval alone barely
separates from random on P@5 and is at or below random on rank quality
(MAP$_\text{B}$/nDCG), confirming that embedding recall orders candidates
only weakly. Rerank lifts P@5 by $\sim$11--15 points; the decisive gain comes
from the reason stage. The best configuration --- SimPO text rerank followed
by Base reason on the top-10 --- reaches \textbf{94.1\%} P@5, $+26.8$ points
over rerank alone and $+39.6$ over retrieval.

\begin{table}[H]
\centering
\caption{Set~1 end-to-end results ($n{=}11$ queries, top-50 labeled). Best
P@5 in \textbf{bold}.}
\label{tab:e2e-set1}
\begin{tabular}{lrrr}
\toprule
Configuration & P@5 (\%) & MAP$_\text{B}$@5 & nDCG@5 \\
\midrule
\multicolumn{4}{l}{\emph{Baselines}} \\
\quad Random                    & 54.0 & 0.415 & 0.540 \\
\quad Retrieval only @5         & 54.5 & 0.383 & 0.523 \\
\quad Retrieval only @10 (P@10) & 56.4 & 0.369 & 0.540 \\
\midrule
\multicolumn{4}{l}{\emph{+ Rerank (top-5 of 50)}} \\
\quad Base $\log p(\texttt{Yes})$, text  & 65.5 & 0.544 & 0.662 \\
\quad SimPO $\log p(\texttt{Yes})$, text & 67.3 & 0.568 & 0.681 \\
\quad SimPO $\log p(\texttt{Yes})$, vision & 69.1 & 0.578 & 0.679 \\
\midrule
\multicolumn{4}{l}{\emph{Reason on all 50 (no rerank)}} \\
\quad Base       & 80.0 & 0.758 & 0.822 \\
\quad SFT$+$GRPO  & 78.2 & 0.731 & 0.807 \\
\quad Kimi-K2.6 (think)  & 78.2 & 0.726 & 0.801 \\
\quad Kimi-K2.6 (no think) & 76.4 & 0.701 & 0.773 \\
\midrule
\multicolumn{4}{l}{\emph{SimPO text rerank $\rightarrow$ reason (top-10)}} \\
\quad Base       & \textbf{94.1} & \textbf{0.860} & \textbf{0.911} \\
\quad Kimi-K2.6 (think)   & 91.8 & 0.796 & 0.866 \\
\quad Kimi-K2.6 (no think) & 89.5 & 0.787 & 0.856 \\
\quad SFT$+$GRPO & 85.9 & 0.717 & 0.796 \\
\midrule
\multicolumn{4}{l}{\emph{Base text rerank $\rightarrow$ reason (top-10)}} \\
\quad Base       & 86.4 & 0.756 & 0.846 \\
\quad SFT$+$GRPO & 83.6 & 0.747 & 0.840 \\
\bottomrule
\end{tabular}
\end{table}

Three observations stand out. First, rerank-then-reason beats
reason-on-all-50 on \emph{both} accuracy and cost: filtering removes weak
candidates that would otherwise dilute the reason ranking, so the smaller
pool is also the better one (94.1 vs.\ 80.0 P@5). Second, the upstream
reranker matters: pairing reason with the SimPO reranker rather than Base
adds $\sim$2--8 points to final P@5, because reason inherits a cleaner
top-10. Third, Base reason outperforms SFT$+$GRPO everywhere under the
deployed prompt --- the offline--end-to-end inversion analyzed in
\S\ref{sec:discussion}.

\paragraph{Set~2.}
Table~\ref{tab:e2e-set2} replicates the headline trend on the larger query
set ($n{=}31$, top-10 labeled). Rerank adds $+6.7$ points over retrieval, and
reason adds a further $+7.6$ to $+11.9$ depending on the reasoner.
SFT$+$GRPO is the weakest reasoner in both sets, but Base and Kimi-K2.6 swap
the top rank: Base $>$ Kimi-K2.6 on Set~1 (94.1 vs.\ 91.8), whereas
Kimi-K2.6 $>$ Base here (91.9 vs.\ 89.3). Gaps also compress on this smaller
labeled pool.

\begin{table}[H]
\centering
\caption{Set~2 end-to-end results ($n{=}31$ queries, top-10 labeled). All
reason rows use the SimPO text reranker upstream. Best P@5 in \textbf{bold}.}
\label{tab:e2e-set2}
\begin{tabular}{lrrr}
\toprule
Configuration & P@5 (\%) & MAP$_\text{B}$@5 & nDCG@5 \\
\midrule
Random                & 68.2 & 0.502 & 0.610 \\
Retrieval only @5     & 73.3 & 0.538 & 0.648 \\
+ SimPO rerank (text) & 80.0 & 0.603 & 0.713 \\
\midrule
\quad $\rightarrow$ Kimi-K2.6 (think)   & \textbf{91.9} & \textbf{0.776} & \textbf{0.850} \\
\quad $\rightarrow$ Kimi-K2.6 (no think) & 91.1 & 0.755 & 0.835 \\
\quad $\rightarrow$ Base       & 89.3 & 0.730 & 0.809 \\
\quad $\rightarrow$ SFT$+$GRPO & 87.6 & 0.695 & 0.783 \\
\bottomrule
\end{tabular}
\end{table}

\subsection{Cost vs.\ Accuracy}
\label{sec:results-cost}

\begin{table}[H]
\centering
\caption{Per-query cost on Set~1 ($n{=}11$). Pareto-optimal rows in
\textbf{bold}.}
\label{tab:e2e-cost}
\begin{tabular}{lrrrrr}
\toprule
Configuration & P@5 (\%) & In & Out & Total & Wall (s) \\
\midrule
\multicolumn{6}{l}{\emph{Rerank only}} \\
\quad SimPO text   & \textbf{67.3} & 251{,}525 & 100 & \textbf{251{,}625} & 12.1 \\
\quad SimPO vision & \textbf{69.1} & 300{,}360 & 100 & \textbf{300{,}460} & 40.4 \\
\quad Base text    & 65.5 & 251{,}525 & 100 & 251{,}625 & 13.6 \\
\midrule
\multicolumn{6}{l}{\emph{Reason on all 50 (no rerank)}} \\
\quad Base       & 80.0 & 782{,}186 & 15{,}127 & 797{,}313 & 29.6 \\
\quad SFT$+$GRPO & 78.2 & 782{,}186 & 11{,}137 & 793{,}324 & 34.1 \\
\quad Kimi-K2.6 (think)   & 78.2 & 1{,}069{,}004 & 482{,}549 & 1{,}551{,}553 & 226.6 \\
\quad Kimi-K2.6 (no think) & 76.4 & 1{,}069{,}144 & 90{,}256 & 1{,}159{,}400 & 148.1 \\
\midrule
\multicolumn{6}{l}{\emph{SimPO text rerank $\rightarrow$ reason (top-10)}} \\
\quad Base       & \textbf{94.1} & 416{,}926 & 2{,}974 & \textbf{419{,}900} & 20.2 \\
\quad SFT$+$GRPO & 85.9 & 416{,}926 & 2{,}341 & 419{,}267 & 21.4 \\
\quad Kimi-K2.6 (think)   & 91.8 & 475{,}594 & 102{,}862 & 578{,}456 & 81.0 \\
\quad Kimi-K2.6 (no think) & 89.5 & 475{,}622 & 19{,}457 & 495{,}079 & 90.4 \\
\midrule
\multicolumn{6}{l}{\emph{Base rerank $\rightarrow$ reason (top-10)}} \\
\quad Base       & 86.4 & 416{,}926 & 2{,}974 & 419{,}900 & 21.8 \\
\quad SFT$+$GRPO & 83.6 & 416{,}926 & 2{,}341 & 419{,}267 & 23.0 \\
\bottomrule
\end{tabular}
\end{table}

Table~\ref{tab:e2e-cost} reports per-query token consumption (Set~1,
$n{=}11$); token columns are average tokens per query. Cost is measured in
tokens (input $+$ output), which is
concurrency-independent and therefore the fair axis; wall-clock is
indicative only, as serving concurrency differs across models (single
A100-SXM4-80GB; Kimi-K2.6 at concurrency 16 on top-10 / 30 on all-50, all
others at 50). Vision rerank excludes a one-time $\sim$39\,s/query image
fetch ($\sim$452 thumbnails).

The Pareto frontier comprises four points, in increasing cost: SimPO text
rerank (67.3\% at 251.6k tokens), SimPO vision rerank (69.1\% at 300.5k),
SimPO rerank $\rightarrow$ SFT$+$GRPO reason (85.9\% at 419.3k), and SimPO
rerank $\rightarrow$ Base reason (94.1\% at 419.9k) --- the clear knee of the
curve. The SFT$+$GRPO point is only marginally non-dominated: 0.6k tokens
cheaper than the Base-reason point but 8.2 points lower, so Base reason is
the practical operating choice. Three trade-offs follow:
\begin{itemize}
  \item \textbf{Reason-on-all-50} costs $\sim$1.9$\times$ the tokens of the
  two-stage pipeline (797k vs.\ 420k) yet scores 14 points lower (80.0 vs.\
  94.1) --- rerank-then-reason is both cheaper and better. The gap is
  starker for Kimi-K2.6: reasoning over all 50 costs $\sim$2.7$\times$ its
  two-stage tokens (1{,}552k vs.\ 578k) for 13.6 points less (78.2 vs.\
  91.8).
  \item \textbf{Kimi-K2.6} emits $\sim$35$\times$ more output tokens (102.9k
  vs.\ 3.0k per query) for \emph{lower} Set~1 accuracy (91.8 vs.\ 94.1),
  making it strictly cost-dominated by the local 4B Base reasoner on this
  set.
  \item \textbf{Vision rerank}, though on the frontier, buys only $+1.8$
  points over text for $+19\%$ input tokens and $\sim$3.3$\times$ the
  latency, so text rerank remains the cost-effective default.
\end{itemize}
\section{Discussion}
\label{sec:discussion}
\label{sec:disc-inversion}

The best offline reason scorer is the weakest fine-tuned reasoner end to end:
the SFT$+$GRPO checkpoint tops the per-criterion leaderboard (59.0 wf1 with
thinking, Table~\ref{tab:reason-leaderboard}) yet trails the untuned Base in
the deployed cascade (85.9 vs.\ 94.1 P@5 on Set~1; 87.6 vs.\ 89.3 on Set~2).
We attribute this inversion primarily to how the labeling task was designed,
through three compounding choices.

\begin{enumerate}
  \item \textbf{Absolute scoring is under-specified.} A single $\{0,1,2\}$
  per-criterion judgment is intrinsically hard, and no scoring rubric
  defining the class boundaries was provided
  (\S\ref{sec:human-annotation}), so each annotator adopted a private
  standard --- a known driver of human label variation \citep{plank2022label}. The evidence audit quantifies the consequence: only 15.9\% of
  score-2 (``match'') labels are backed by verbatim profile evidence
  (Table~\ref{tab:evidence-audit}).
  \item \textbf{The interface grouped items by criteria set.} Annotators
  scored all candidates under one brief consecutively
  (\S\ref{sec:human-annotation}), so within-batch labels are correlated
  rather than independent judgments.
  \item \textbf{The two regimes use different human rubrics.} End-to-end
  relevance is a holistic KOL-level pass/fail, whereas the reason scorer is
  supervised one criterion at a time and aggregated by an even sum. Even
  with clean labels, a model fit to per-criterion targets optimizes an
  objective that only loosely tracks the holistic signal --- in our reading,
  the primary cause of the inversion. Choices 1--2 compound this: the
  per-criterion targets are not only a proxy objective but a noisy one.
\end{enumerate}

\paragraph{The plateau is label-imposed, not capacity-imposed.}
Restricting evaluation to the $\approx$38\% of test items the evidence audit
marks \emph{clear-cut}, the SFT checkpoint rises from 57.7 to \textbf{73.4}
wf1 (clear-2 F1 91.7) and SFT$+$GRPO reaches \textbf{80.6}.
On clean labels the 4B model is excellent; the full-set plateau reflects the
model \emph{correctly disagreeing} with ungrounded labels. This corroborates
choice~1: the fine-tuned scorer learned the annotators' generous, weakly
grounded ``yes'' --- exactly the signal that does not survive the holistic
end-to-end test.

\paragraph{Text rerank over vision.}
The deployed reranker is text-only. Adding thumbnails carries a real serving
cost --- ${\sim}19\%$ more input tokens, ${\sim}3.3\times$ the rerank
wall-clock, plus a one-time image fetch --- while buying only $+1.8$ P@5
end-to-end (Table~\ref{tab:e2e-set1}). The marginal gain does not justify
the added latency and image-processing cost.

\paragraph{Why filtering before reasoning wins.}
Reasoning over all 50 candidates loses to rerank-then-reason (80.0 vs.\ 94.1
P@5) despite seeing strictly more of the pool. We read the gap as a division
of labor between complementary stages. The reranker, trained on pairwise
preferences, is strong at \emph{relative} filtering but is not a precise
judge --- rerank alone tops out at 67--69 P@5. The reason scorer is the
precise judge, but its per-criterion even sum has coarse resolution over a
large pool: with five criteria scored in $\{0,1,2\}$, a 50-candidate field
produces ties and near-ties in which weak candidates dilute the top-5.
Filtering first with the preference-trained reranker hands the reasoner a
short pool where its precision is decisive; each stage supplies what the
other lacks.
\section*{Limitations}
\label{sec:limitations}

\paragraph{Small samples.}
Set~1 has $n{=}11$ queries and Set~2 $n{=}31$; the pairwise test set has 41
triplets, so a single flipped triplet moves EM by $\approx$2.4pp.
Differences of a few points should be read as directional, not significant.

\paragraph{Audit provenance.}
The evidence audit is produced by a frontier model acting as an independent
auditor; although every quote is verified as an exact substring of the
profile (a model-independent check), the selection of quotes and the
grounding judgment inherit the auditor's biases.

\paragraph{Synthetic briefs and single market.}
Training and evaluation queries derive from synthetic briefs; despite
schema-mirroring and web grounding, their distribution may differ from real
proprietary briefs. All experiments target the Thai KOL market with one
model family (4B), so transfer to other languages, markets, and scales is
untested.

\paragraph{Serving comparability.}
Wall-clock timings are not like-for-like across models (vLLM stages at
concurrency 50, the frontier baseline at 16--30); cost conclusions rest on
the token axis.
\section{Conclusion}
\label{sec:conclusion}

We presented InfluMatch, a three-stage retrieval $\rightarrow$ rerank
$\rightarrow$ reason cascade that matches free-form Thai marketer criteria
to KOLs using only small open-weight models. The deployed system --- a
SimPO-tuned 4B reranker followed by an untuned 4B reason scorer --- reaches
89--94\% P@5 against human relevance labels, matching or approaching a
frontier model at ${\sim}35\times$ fewer output tokens, and shows that the
cascade design itself, filtering before reasoning, is both cheaper and more
accurate than reasoning over the full candidate pool.

Beyond the system, our central finding concerns supervision. Of the two
fine-tuned students, only the one trained on \emph{relative} judgments
survived end-to-end evaluation: pairwise SimPO transferred and matched the
frontier baseline, while pointwise SFT$+$GRPO topped the offline leaderboard
yet degraded the deployed ranking --- an inversion we trace, via an evidence
audit, to the design of the absolute labeling task. On this evidence,
pointwise fine-tuning was the wrong objective \emph{for this label
distribution}: the recoverable supervision signal in
this domain lives in relative judgments, not absolute scores.

\bibliographystyle{unsrtnat-etal}
\bibliography{references} 

\end{document}